\documentclass[11pt,a4paper]{article}
 
\usepackage[hyperref]{naaclhlt2018}
\usepackage{times}
\usepackage{latexsym}
\usepackage{amsmath}
\usepackage{url}
\usepackage{graphicx}

\newcommand{\mvec}[1]{\vec{\mathstrut \smash[t]{#1}}}

\aclfinalcopy
\def\aclpaperid{3017}

\title{\textbf{Metric for Automatic Machine Translation Evaluation \\ based on Universal Sentence Representations}}

\author{Hiroki Shimanaka$^{\dagger}$ \\\And
  Tomoyuki Kajiwara$^{\dagger\ddagger}$ \\
  $^{\dagger}$Graduate School of Systems Design, Tokyo Metropolitan University, Tokyo, Japan \\
  {\tt shimanaka-hiroki@ed.tmu.ac.jp, komachi@tmu.ac.jp} \\
  $^{\ddagger}$Institute for Datability Science, Osaka University, Osaka, Japan \\
  {\tt kajiwara@ids.osaka-u.ac.jp} \\\And
  Mamoru Komachi$^{\dagger}$ \\
  }

\date{}

\begin{document}
\maketitle

\begin{abstract}
Sentence representations can capture a wide range of information that cannot be captured by local features based on character or word N-grams.
This paper examines the usefulness of universal sentence representations for evaluating the quality of machine translation.
Although it is difficult to train sentence representations using small-scale translation datasets with manual evaluation, sentence representations trained from large-scale data in other tasks can improve the automatic evaluation of machine translation.
Experimental results of the WMT-2016 dataset show that the proposed method achieves state-of-the-art performance with sentence representation features only.
\end{abstract}

\section{Introduction}\label{sec:intro}
This paper describes a segment-level metric for automatic machine translation evaluation (MTE).
MTE metrics having a high correlation with human evaluation enable the continuous integration and deployment of a machine translation (MT) system.
Various MTE metrics have been proposed in the metrics task of the Workshops on Statistical Machine Translation (WMT) that was started in 2008.
However, most MTE metrics are obtained by computing the similarity between an MT hypothesis and a reference translation based on character N-grams or word N-grams, such as SentBLEU~\cite{lin-2004}, which is a smoothed version of BLEU~\cite{papineni-2002}, Blend~\cite{ma-2017}, MEANT~2.0~\cite{lo-2017}, and chrF++~\cite{popovic-2017}, which achieved excellent results in the WMT-2017 Metrics task~\cite{bojar-2017}.
Therefore, they can exploit only limited information for segment-level MTE.
In other words, MTE metrics based on character N-grams or word N-grams cannot make full use of sentence representations; they only check for word matches.

We propose a segment-level MTE metric by using universal sentence representations capable of capturing information that cannot be captured by local features based on character or word N-grams.
The results of an experiment in segment-level MTE conducted using the datasets for to-English language pairs on WMT-2016 indicated that the proposed regression model using sentence representations achieves the best performance.

The main contributions of the study are summarized below:
\begin{itemize}
\item We propose a novel supervised regression model for segment-level MTE based on universal sentence representations.
\item We achieved state-of-the-art performance on the WMT-2016 dataset for to-English language pairs without using any complex features and models. 
\end{itemize}

\begin{figure*}[t]
  \begin{minipage}{0.33\hsize}
    \begin{center}
      \includegraphics[scale=0.28]{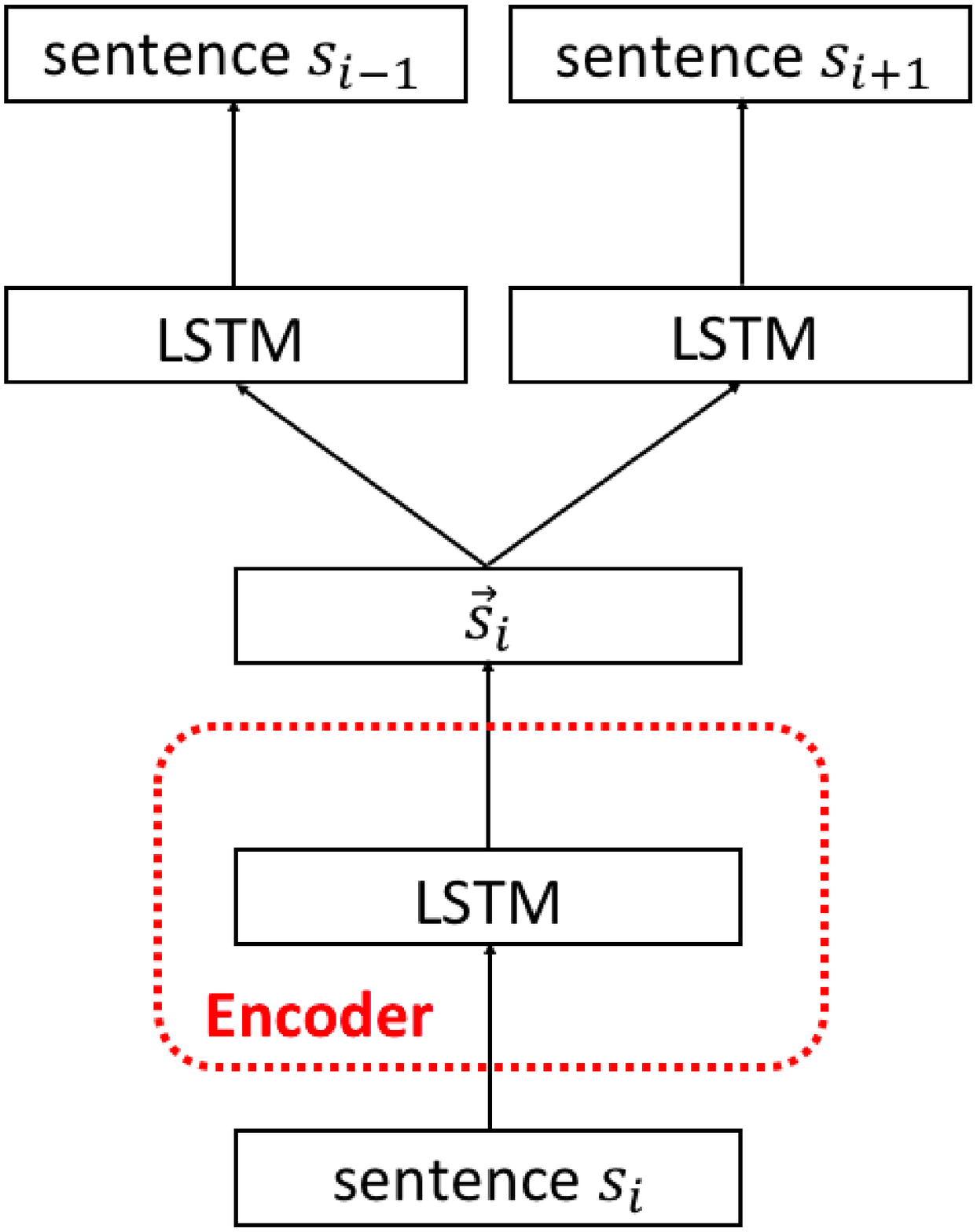}
      \caption{Outline of Skip-Thought.}\label{fig:skip-thought}
    \end{center}
  \end{minipage}
  \begin{minipage}{0.33\hsize}
    \begin{center}
      \includegraphics[scale=0.28]{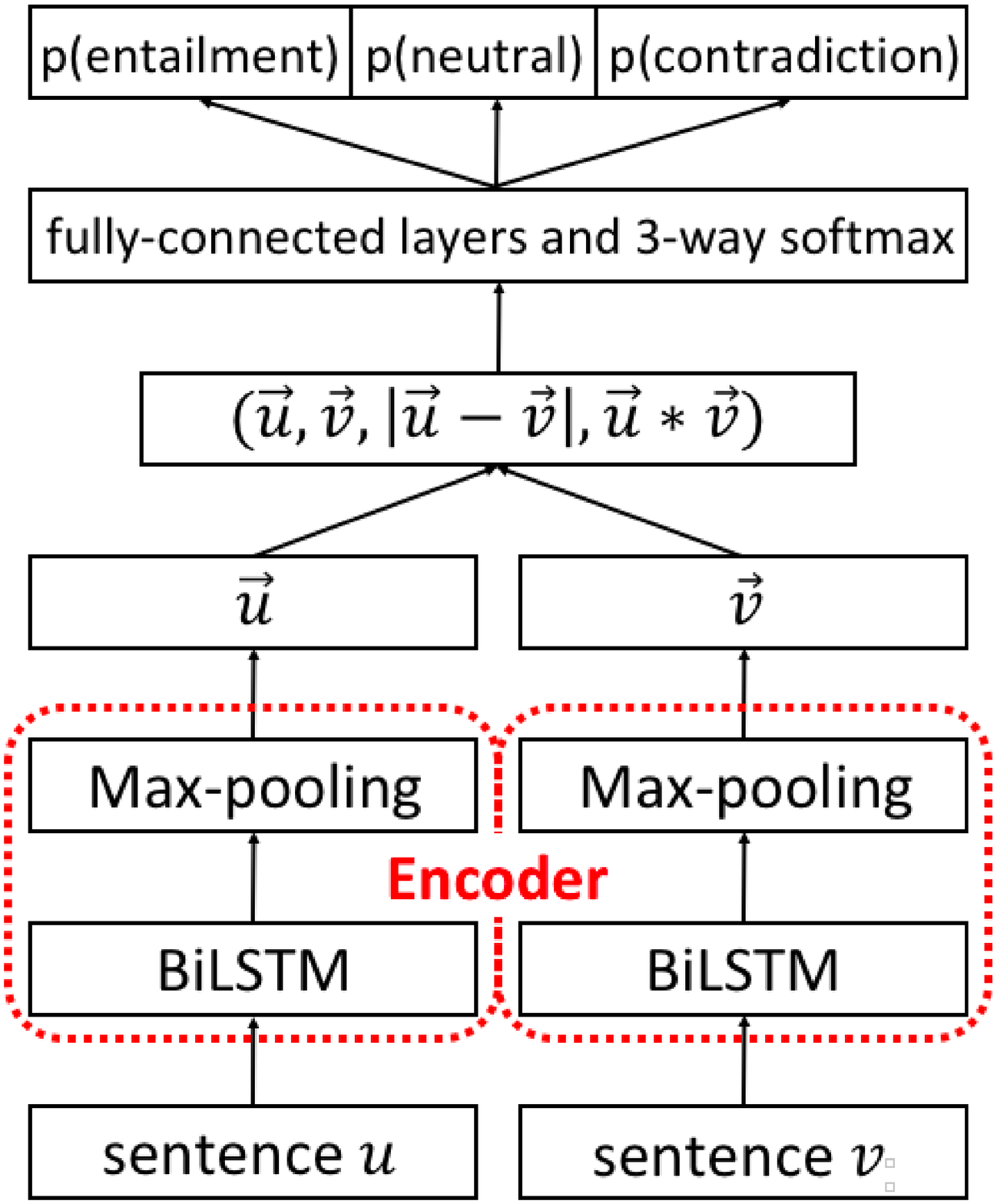}
      \caption{Outline of InferSent.}\label{fig:infersent}
    \end{center}
  \end{minipage}
  \begin{minipage}{0.33\hsize}
    \begin{center}
      \includegraphics[scale=0.28]{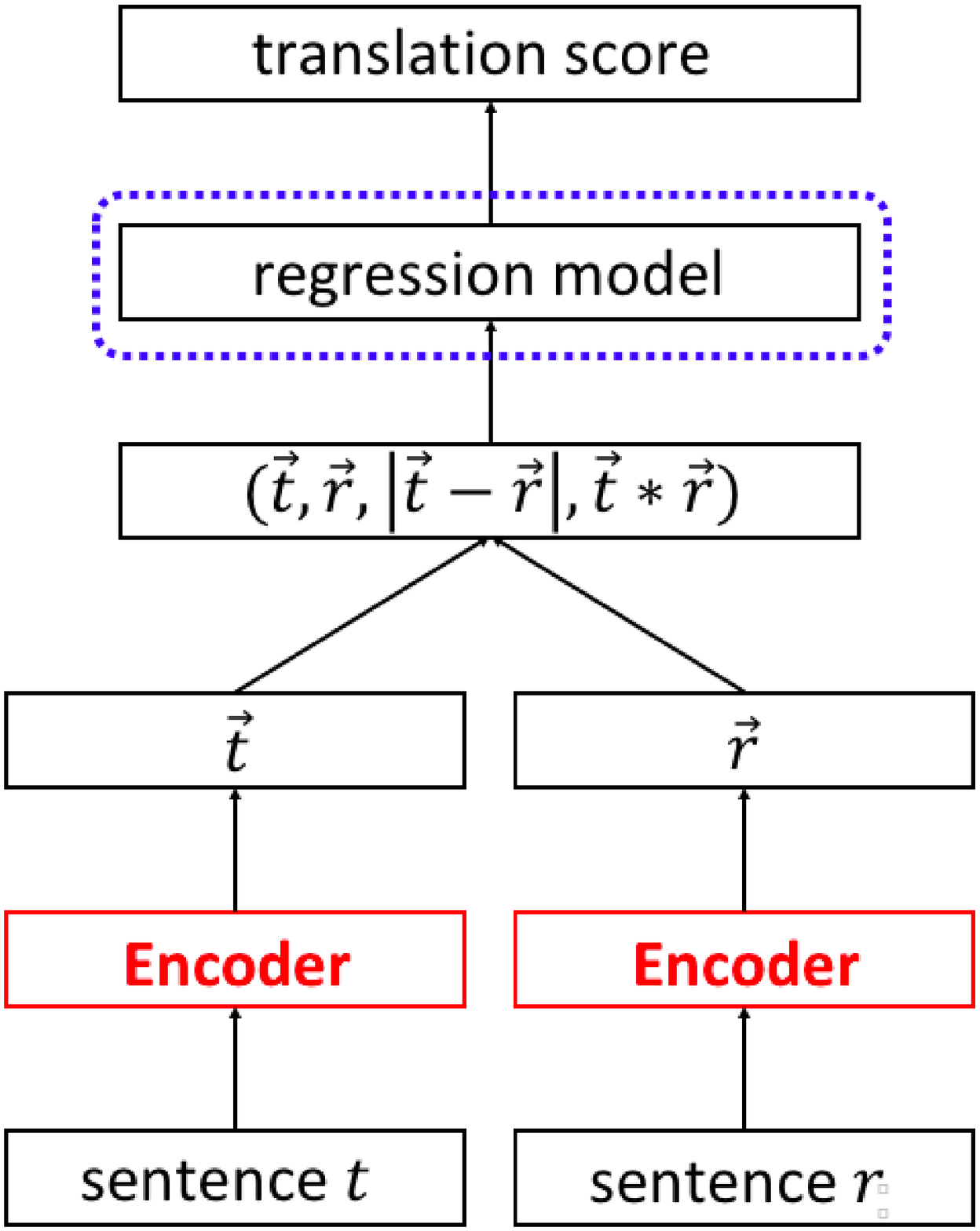}
      \caption{Outline of our metric.}\label{fig:regression}
    \end{center}
  \end{minipage}
\end{figure*}

\section{Related Work}\label{sec:related}
DPMF$_{\mathrm{comb}}$~\cite{yu-2015a} achieved the best performance in the WMT-2016 Metrics task~\cite{bojar-2016}.
It incorporates 55 default metrics provided by the Asiya MT evaluation toolkit\footnote{http:\slash{}\slash{}asiya.lsi.upc.edu\slash{}}~\cite{PBML_Asiya:2010}, as well as three other metrics, namely, DPMF~\cite{yu-2015b}, REDp~\cite{yu-2015a}, and ENTFp~\cite{yu-2015a}, using ranking SVM to train parameters of each metric score.
DPMF evaluates the syntactic similarity between an MT hypothesis and a reference translation.
REDp evaluates an MT hypothesis based on the dependency tree of the reference translation that comprises both lexical and syntactic information.
ENTFp~\cite{yu-2015a} evaluates the fluency of an MT hypothesis.

After the success of DPMF$_{\mathrm{comb}}$, Blend\footnote{http:\slash{}\slash{}github.com\slash{}qingsongma\slash{}blend}~\cite{ma-2017} achieved the best performance in the WMT-2017 Metrics task~\cite{bojar-2017}.
Similar to DPMF$_{\mathrm{comb}}$, Blend is essentially an SVR (RBF kernel) model that uses the scores of various metrics as features.
It incorporates 25 lexical metrics provided by the Asiya MT evaluation toolkit, as well as four other metrics, namely, BEER~\cite{stanojevic-2015a}, CharacTER~\cite{wang-2016}, DPMF and ENTFp.
BEER~\cite{stanojevic-2015a} is a linear model based on character N-grams and replacement trees.
CharacTER~\cite{wang-2016} evaluates an MT hypothesis based on character-level edit distance.

DPMF$_{\mathrm{comb}}$ is trained through relative ranking of human evaluation data in terms of relative ranking (RR). The quality of five MT hypotheses of the same source segment are ranked from 1 to 5 via comparison with the reference translation.
In contrast, Blend is trained through direct assessment (DA) of human evaluation data. DA provides the absolute quality scores of hypotheses, by measuring to what extent a hypothesis adequately expresses the meaning of the reference translation.
The results of the experiments in segment-level MTE conducted using the datasets for to-English language pairs on WMT-2016 showed that Blend achieved a better performance than DPMF$_{\mathrm{comb}}$ (Table~\ref{tab:experiment}).
In this study, as with Blend, we propose a supervised regression model trained using DA human evaluation data.

Instead of using local and lexical features, ReVal\footnote{https:\slash{}\slash{}github.com\slash{}rohitguptacs\slash{}ReVal}~\cite{gupta-2015a, gupta-2015b} proposes using sentence-level features.
It is a metric using Tree-LSTM (Tai et al., 2015) for training and capturing the holistic information of sentences.
It is trained using datasets of pseudo similarity scores, which is generated by translating RR data, and out-domain datasets of similarity scores of SICK\footnote{http:\slash{}\slash{}clic.cimec.unitn.it\slash{}composes\slash{}sick.html}.
However, the training dataset used in this metric consists of approximately 21,000 sentences; thus, the learning of Tree-LSTM is unstable and accurate learning is difficult (Table~\ref{tab:experiment}).
The proposed metric uses sentence representations trained using LSTM as sentence information.
Further, we apply universal sentence representations to this task; these representations were trained using large-scale data obtained in other tasks.
Therefore, the proposed approach avoids the problem of using a small dataset for training sentence representations.

\section{Regression Model for MTE Using Universal Sentence Representations}\label{sec:proposed_method}
The proposed metric evaluates MT results with universal sentence representations trained using large-scale data obtained in other tasks.
First, we explain two types of sentence representations used in the proposed metric in Section~\ref{sub:sentence_embedding}.
Then, we explain the proposed regression model and feature extraction for MTE in Section~\ref{sub:regression_model}.

\begin{table*}[t]
\centering
\begin{tabular}{c|c|c|c|c|c|c}
         & cs-en & de-en & fi-en & ro-en & ru-en & tr-en \\ \hline
WMT-2015 & 500   & 500   & 500   & -     & 500   & -     \\ \hline
WMT-2016 & 560   & 560   & 560   & 560   & 560   & 560   \\
\end{tabular}
\caption{Number of DA human evaluation datasets for to-English language pairs\protect \footnotemark[8] in WMT-2015~\cite{stanojevic-2015b} and WMT-2016~\cite{bojar-2016}.}
\label{tab:dataset}
\end{table*}

\subsection{Universal Sentence Representations}\label{sub:sentence_embedding}
Several approaches have been proposed to learn sentence representations. These sentence representations are learned through large-scale data so that they constitute potentially useful features for MTE.
These have been proved effective in various NLP tasks such as document classification and measurement of semantic textual similarity, and we call them universal sentence representations.

First, Skip-Thought\footnote{https:\slash{}\slash{}github.com\slash{}ryankiros\slash{}skip-thoughts}~\cite{kiros-2015} builds an unsupervised model of universal sentence representations trained using three consecutive sentences, such as $s_{i-1}$, $s_i$, and $s_{i+1}$.
It is an encoder-decoder model that encodes sentence $s_i$ and predicts previous and next sentences $s_{i-1}$ and $s_{i+1}$ from its sentence representation $\vec{s_i}$ (Figure~\ref{fig:skip-thought}).
As a result of training, this encoder can produce sentence representations.
Skip-Thought demonstrates high performance, especially when applied to document classification tasks.

Second, InferSent\footnote{https:\slash{}\slash{}github.com\slash{}facebookresearch\slash{}InferSent}~\cite{conneau-2017} constructs a supervised model computing universal sentence representations trained using Stanford Natural Language Inference (SNLI) datasets\footnote{https:\slash{}\slash{}nlp.stanford.edu\slash{}projects\slash{}snli\slash{}}~\cite{bowman-2015}.
The Natural Language Inference task is a classification task of sentence pairs with three labels, {\it entailment}, {\it contradiction} and {\it neutral}; thus, InferSent can train sentence representations that are sensitive to differences in meaning.
This model encodes sentence pairs $u$ and $v$ and generates features by sentence representations $\vec{u}$ and $\vec{v}$ with a bi-directional LSTM architecture with max pooling (Figure~\ref{fig:infersent}).
InferSent demonstrates high performance across various document classification and semantic textual similarity tasks.

\subsection{Regression Model for MTE}\label{sub:regression_model}
In this paper, we propose a segment-level MTE metric for to-English language pairs.
This problem can be treated as a regression problem that estimates translation quality as a real number from an MT hypothesis $t$ and a reference translation $r$.
Once $d$-dimensional sentence vectors $\vec{t}$ and $\vec{r}$ are generated, the proposed model applies the following three matching methods to extract relations between $t$ and $r$ (Figure~\ref{fig:regression}).

\begin{itemize}
  \item Concatenation: $(\mvec{t}, \mvec{r})$
  \item Element-wise product: $\mvec{t} * \mvec{r}$
  \item Absolute element-wise difference: $|\mvec{t} - \mvec{r}|$
\end{itemize}

Thus, we perform regression using $4d$-dimensional features of $\mvec{t}$, $\mvec{r}$, $\mvec{t} * \mvec{r}$ and $|\mvec{t} - \mvec{r}|$.

%The proposed metric only trains regression models and does not fine-tune sentence representations.
%We also performed experiments using multiple sentence encoders.
%In this experiment, the features of the regression model are obtained by concatenating the aforementioned $4d$ dimension feature vectors generated from each encoder.

\section{Experiments of Segment-Level MTE for To-English Language Pairs}\label{sec:experiment}
We performed experiments using evaluation datasets of the WMT Metrics task to verify the performance of the proposed metric.

\begin{table*}[t]
\centering
\begin{tabular}{l|cccccc|c}
                                         &      cs-en  &      de-en  &      fi-en  &      ro-en  &      ru-en  &      tr-en  &      Avg.   \\ \hline
SentBLEU~\cite{bojar-2016}               &      0.557  &      0.448  &      0.484  &      0.499  &      0.502  &      0.532  &      0.504  \\ \hline
Blend~\cite{ma-2017}                     &      0.709  &      0.601  &      0.584  &      0.636  &      0.633  & {\bf 0.675} &      0.640  \\
DPMF$_{\mathrm{comb}}$~\cite{bojar-2016} & {\bf 0.713} &      0.584  &      0.598  &      0.627  &      0.615  &      0.663  &      0.633  \\
ReVal~\cite{bojar-2016}                  &      0.577  &      0.528  &      0.471  &      0.547  &      0.528  &      0.531  &      0.530  \\ \hline
SVR with Skip-Thought                    &      0.665  &      0.571  &      0.609  & {\bf 0.677} &      0.608  &      0.599  &      0.622  \\
SVR with InferSent                       &      0.679  &      0.604  &      0.617  &      0.640  &      0.644  &      0.630  &      0.636  \\
SVR with InferSent + Skip-Thought        &      0.686  & {\bf 0.611} & {\bf 0.633} &      0.660  & {\bf 0.649} &      0.646  & {\bf 0.648} \\
\end{tabular}
\caption{Segment-level Pearson correlation of metric scores and DA human evaluations scores for to-English language pairs in WMT-2016 (newstest2016).}
\label{tab:experiment}
\end{table*}

\subsection{Setups}\label{sub:setting}
\paragraph{Datasets.}We used datasets for to-English language pairs from the WMT-2016 Metrics task~\cite{bojar-2016} as summarized in Table~\ref{tab:dataset}.
Following Ma et al.~\shortcite{ma-2017}, we employed all other to-English DA data as training data (4,800 sentences) for testing on each to-English language pair (560 sentences) in WMT-2016.
%For the evaluation, we used datasets for to-English language pairs used in the WMT-2016 Metrics task~\cite{bojar-2016} (newstest2016).
%For training the regression model, we used only 4,800 instances extracted from sampled DA human evaluation data of newstest2015 in WMT-2015~\cite{stanojevic-2015b} and newstest2016 in WMT-2016 for each to-English language pair.

\footnotetext[8]{en: English, cs: Czech, de: German, fi: Finnish, ro: Romanian, ru: Russian, tr: Turkish}
\addtocounter{footnote}{1}

\paragraph{Features.}Publicly available pre-trained sentence representations such as Skip-Thought\footnotemark[5] and InferSent\footnotemark[6] were used as the features mentioned in Section~\ref{sec:proposed_method}.
Skip-Thought is a collection of 4,800-dimensional sentence representations trained on 74 million sentences of the BookCorpus dataset~\cite{zhu-2015}.
InferSent is a collection of 4,096-dimensional sentence representations trained on both 560,000 sentences of the SNLI dataset~\cite{bowman-2015} and 433,000 sentences of the MultiNLI dataset~\cite{williams-2017}.

\paragraph{Model.}Our regression model used SVR with the RBF kernel from scikit-learn\footnote{http:\slash{}\slash{}scikit-learn.org\slash{}stable\slash{}}.
Hyper-parameters were determined through 10-fold cross validation using the training data.
We examined all combinations of hyper-parameters among $C \in \{0.01, 0.1, 1.0, 10\}$, $\epsilon \in \{0.01, 0.1, 1.0, 10\}$, and $\gamma \in \{0.01, 0.1, 1.0, 10\}$.
%We estimated each parameter of the SVR model by using grid search with cross-validation: $C \in \{0.01, 0.1, 1.0, 10\}$, $\epsilon \in \{0.01, 0.1, 1.0, 10\}$, and $\gamma \in \{0.01, 0.1, 1.0, 10\}$.

There are three comparison methods: Blend~\cite{ma-2017}, DPMF$_{\mathrm{comb}}$~\cite{yu-2015a}, and ReVal~\cite{gupta-2015a, gupta-2015b}, as described in Section~\ref{sec:related}.
Blend and DPMF$_{\mathrm{comb}}$ are MTE metrics that exhibited the best performance in the WMT-2017 Metrics task~\cite{bojar-2017} and WMT-2016 Metrics task, respectively.
We compared the Pearson correlation of each metric score and DA human evaluation scores.

\subsection{Result}\label{sub:result}
%The experimental results are listed in Table~\ref{tab:experiment}.
As can be seen in Table~\ref{tab:experiment}, the proposed metric, which combines InferSent and Skip-Thought representations, surpasses the best performance in three out of six to-English languages pairs and achieves state-of-the-art performance on average.

\subsection{Discussion}\label{sub:discussion}
These results indicate that it is possible to adopt universal sentence representations in MTE by training a regression model using DA human evaluation data.
%As the proposed metric shows a better performance than Blend, which ensembles scores of various metrics as features, it seems that it is more important to exploit large-scale data to pre-train universal sentence representations instead of combining various metrics learned from small or limited data.
%ReVal cannot show high correlation of metric scores with DA human evaluation scores.
%This result shows that the larger the data used to pre-train a regression model, the more effective the sentence representation for MTE.
Since Blend is an ensemble method using combinations of various MTE metrics as features, our results show that universal sentence representations can consider information more abundantly than a complex model.
Since ReVal is also based on sentence representations, we conclude that universal sentence representations trained on a large-scale dataset are more effective for MTE tasks than sentence representations trained on a small or limited in-domain dataset.

\subsection{Error Analysis}\label{sub:analysis}
We re-implemented Blend\footnote{http:\slash{}\slash{}github.com\slash{}qingsongma\slash{}blend}~\cite{ma-2017} and compared the evaluation results with the proposed metric.\footnote{The average Pearson correlation of all language pairs after re-implementing Blend was 0.636, which is a little lower than the value reported in their paper. However, we judged that the following discussion will not be affected by this difference.}

%We analyzed the top 112 MT hypotheses with high DA human scores. In other words, the top 112 MT hypotheses that were close to the meaning of the reference translations for each to-English language pairs were analyzed.
%A total of 70 MT hypotheses for all to-English language pairs were correctly evaluated in Blend, whereas 88 MT hypotheses were correctly evaluated in the proposed metric.
We analyzed 20\% of the pairs of MT hypotheses and reference translations (112 sentence pairs $\times$ 6 languages = 672 sentence pairs) in descending order of DA human score in each language pair.
In other words, the top 20\% of MT hypotheses that were close to the meaning of the reference translations for each language pair were analyzed.
Among these, only Blend estimates the translation quality as high for 70 sentence pairs, and only our metric estimates the translation quality as high for 88 sentence pairs.
%Here, we ranked each pair by the estimated value of translation quality in each method, and defined the top 20\% as high quality.

\paragraph{Surface.}
%There were 26 MT hypotheses with a low matching rate of surface with the reference translations out of the 70 MT hypotheses that were correctly evaluated in Blend. However, there were 42 such MT hypotheses out of 88 MT hypotheses that were correctly evaluated in the proposed metric.
Among pairs estimated to have high translation quality by each method, there were 26 pairs in Blend and 42 pairs in the proposed method with a low word surface matching rate between MT hypotheses and reference translations.
This result shows that the proposed metric can evaluate a wide range of sentence information that cannot be captured by Blend.

%\paragraph{Preprocessing}There were 15 MT hypotheses containing words with the same meaning but different surface out of the 70 MT hypotheses that were correctly evaluated in Blend. However, there were only two MT hypotheses out of the 88 MT hypotheses that were correctly evaluated in the proposed metric.
%This is because Blend evaluates sentences in lower-case, but the proposed metric evaluates them in true-case.
%Skip-Thought~\cite{kiros-2015} and InferSent~\cite{conneau-2017} use pre-trained true-cased word representations to generate sentence representations.
%For this reason, in the proposed metric, it seems that sentence representations are generated by words with the same meaning but different surfaces; thus, correct evaluation is not possible.
%It seems that a more accurate evaluation in the proposed metric will be possible by training DA human evaluation scores with sentence representations trained in lower-case.

\paragraph{Unknown words.}
There were 26 MT hypotheses consisting of words that were treated as unknown words in Skip-Thought or InferSent that were correctly evaluated in Blend.
On the other hand, there were 26 MT hypotheses that were correctly evaluated in the proposed metric.
This result shows that the proposed metric is affected by unknown words.
However, it is also true that there are some MT hypotheses containing unknown words that can be correctly evaluated.
Therefore, we analyzed further by focusing on sentence length.
There were 17 MT hypotheses consisting of words that were treated as unknown words by either Skip-Thought or InferSent with a short length (15 words or less) that were correctly evaluated in Blend.
However, in the proposed metric, there were only two MT hypotheses that were correctly evaluated.
This result indicates that the shorter the sentence, the more likely is the proposed metric to be affected by unknown words.

\section{Conclusions}\label{sec:outro}
In this study, we tried to apply universal sentence representation to MTE based on the DA of human evaluation data.
%The experimental results indicated that an MTE metric based on DA by using universal sentence representations has a high correlation with human evaluation.
Our segment-level MTE metric achieved the best performance on the WMT-2016 dataset.
We conclude that:
\begin{itemize}
  \item Universal sentence representations can consider information more comprehensively than an ensemble metric using combinations of various MTE metrics based on features of character or word N-grams.
  \item Universal sentence representations trained on a large-scale dataset are more effective than sentence representations trained on a small or limited in-domain dataset.
  \item Although a metric based on SVR with universal sentence representations is not good at handling unknown words, it correctly estimates the translation quality of MT hypotheses with a low word matching rate with reference translations.
\end{itemize}

%In this study, we only focused on segment-level MTE.
%We were not able to conduct experiments using to-English language pairs at the system level.
%We will try to verify whether our metric is effective for MTE at the system level.
%In addition, the proposed metric is effective if there is an universal sentence representation and DA human evaluation data.
%Thus, we will also try to build a dataset for universal sentence representations in another language and perform an experiment with various languages other than to-English.
Following the success of InferSent~\cite{conneau-2017}, many works~\cite{wieting-2017,cer-2018,subramanian-2018} on universal sentence representations have been published.
Based on the results of our work, we expect that the MTE metric will be further improved using these better universal sentence representations.

\bibliographystyle{acl_natbib}
\bibliography{naacl18}  
\end{document}